\documentclass[preprint,11pt]{elsarticle} 

\usepackage{graphicx}
\usepackage{amssymb}
\usepackage{hyperref}

\usepackage{lscape}
\usepackage{caption}
\usepackage{subcaption}
\usepackage{xcolor}
\usepackage{enumitem}
\usepackage{textcomp}
\usepackage{mathtools}

\usepackage{array}
\usepackage{colortbl}
\usepackage{xcolor}
\usepackage{textpos}
\definecolor{light-gray}{gray}{0.90}
\usepackage{multirow}
\usepackage{tablefootnote}
\usepackage{comment}
\usepackage{amsmath,amsfonts}
\usepackage[margin=0.9in]{geometry}
\usepackage{setspace}

\usepackage{wasysym}

\usepackage[font=small]{caption} 
\newcolumntype{C}[1]{>{\centering\arraybackslash\hspace{0pt} }p{#1} }
\newcolumntype{R}[1]{>{\raggedleft\arraybackslash\hspace{0pt} }p{#1} }
\newcolumntype{L}[1]{>{\raggedright\arraybackslash\hspace{0pt} }p{#1} }
\newcolumntype{P}[1]{>{\hspace{0pt} }p{#1}}

\journal{ARXiV}

\begin{document}
\begin{frontmatter}
\title{Applying Artificial Intelligence for Age Estimation in Digital Forensic Investigations}



\author{Thomas Gr\"ubl, Harjinder Singh Lallie\corref{cor1}}
\ead{thomasgruebl@gmx.at, HL@warwick.ac.uk}
\address{WMG, University of Warwick, Coventry, CV4 7AL}

\cortext[cor1]{Corresponding author}

\begin{abstract}
The precise age estimation of child sexual abuse and exploitation (CSAE) victims is one of the most significant digital forensic challenges. Investigators often need to determine the age of victims by looking at images and interpreting the sexual development stages and other human characteristics. The main priority – safeguarding children – is often negatively impacted by a huge forensic backlog, cognitive bias and the immense psychological stress that this work can entail.

This paper evaluates existing facial image datasets and proposes a new dataset tailored to the needs of similar digital forensic research contributions. This small, diverse dataset of 0 to 20-year-old individuals contains 245 images and is merged with 82 unique images from the FG-NET dataset, thus achieving a total of 327 images with high image diversity and low age range density. The new dataset is tested on the Deep EXpectation (DEX) algorithm pre-trained on the IMDB-WIKI dataset.

The overall results for young adolescents aged 10 to 15 and older adolescents/adults aged 16 to 20 are very encouraging - achieving MAEs as low as 1.79, but also suggest that the accuracy for children aged 0 to 10 needs further work. In order to determine the efficacy of the prototype, valuable input of four digital forensic experts, including two forensic investigators, has been taken into account to improve age estimation results. Further research is required to extend datasets both concerning image density and the equal distribution of factors such as gender and racial diversity.
\end{abstract}


\end{frontmatter}


\section{Introduction}\label{Introduction}

One of the key challenges in the investigation of child sexual abuse and exploitation (CSAE) material relates to the age estimation of both the abused and the abuser. Humans can estimate several facets, including age, height, weight, gender, ethnicity, and facial expressions reasonably easily. However, human decision making in this area can be impacted by the size and volume of images to be processed and also the trauma caused by the process of assessment.

Of equal concern are cognitive biases presented to the investigator. These are often induced by traumatic stress and other psychological disorders caused by the regular examination and processing of this type of evidence  \citep{sanchez2019}. Distress and anxiety, or situations that create feelings of disgust can render impaired decision-making \citep{lerner2004} and negative emotions render risky decisions \citep{heilman2010}. Furthermore, human factors such as cognitive bias can also impede effective decision-making \citep{sunde2019}. The motivation for saving a child can influence rational analyses \citep{sunde2019}. In some cases, staff exposure to disturbing material can lead to immense psychological stress \citep{jones2012}.

Machine learning (ML) techniques can aid the development of Decision Support Systems (DSS). ML can help to increase estimation accuracy but also reduce human exposure time to CSAE material. Algorithms must be trained intensively before they can achieve acceptable estimation accuracy \citep{anda2018}, however, generally, whenever a suitable algorithm is trained with sufficient sample images, it often outperforms the human judgements \citep{anda2018, antipov2016, escalera2016, rothe2015, rothe2016}.

ML techniques can serve as complementary tools for forensic investigators. These systems can aid in identifying, restoring and gathering different kinds of evidence, processing the material via suitable reasoning modules (e.g. AI classifiers), and presenting the results to an investigator for final decision making  \citep{mukkamala2003, irons2014, costantini2019}. A number of ML techniques have been proposed to provide accurate age predictions using facial images \citep{anand2017, antipov2016, chen2017, escalera2016, feng2019, rothe2015, rothe2016}. However, ML techniques are rarely used in day to day investigations. Perhaps one of the most promising techniques in the field of ML are Convolutional Neural Networks (CNN), which are able to make predictions about a particular input based on the training data they have received beforehand. In the case of age estimation in digital forensics, CNNs are trained with facial images of CSAE victims and the corresponding age labels. Depending on the size, quality and diversity of the training dataset, the neural network is able to output accurate age estimation results on completely new testing images.

This study proposes a prototype that receives an input image and classifies the image into an age category. This research will benefit both crime victims as well as digital forensic investigators. Although a lot of further research must be done in this field, the findings act as the first step to faster, more accurate, less biased and less emotionally draining digital forensic investigations.
\section{Background}\label{Background}

Current investigative techniques are steadily becoming obsolete, not only due to advancements in technology but also a growth in cybercrime and limitations in time and resources \citep{irons2014}. To handle these challenges, intelligent tools must be implemented, which help to tackle intricate problems. Researchers have identified a number of important research challenges which include: 

\begin{itemize}[noitemsep]
\item Increasing demand to analyse the overwhelming bulk of evidence and resulting in significant delays in the judicial process \citep{anda2018, casey2009, garfinkel2010, hoelz2009, peersman2016}.
\item Lack of intelligent analytical approaches to investigations \citep{atlam2020, costantini2019, ganesh2017, irons2014, krivchenkov2019, peersman2016}.
\item Staff exposure to disturbing material \citep{bourke2014, powell2015, sanchez2019, seigfried2018}.
\end{itemize}

These problems are long-standing digital forensic challenges. Although acknowledged for a while that the application of ML techniques can yield improvements in digital forensic processes \citep{beebe2009, irons2014}, practical research contributions in this field are still limited and the lack of diverse approaches in the recent literature can be interpreted as a stagnation of the research field and needs to be addressed immediately.

CSAE remains a high-volume offence, with an upsurge in recorded incidents across the UK. Between 2017 and 2018, the number of reports of suspected CSAE material in the UK has risen by over a third \citep{nca2019}. Similarly, the number of global reports to the National Center for Missing and Exploited Children has risen by over 80\%. The UK National Crime Agency reported that in 2019, over 300,000 individuals in the UK alone were estimated to present a threat to children and 7,600 children were safeguarded in relation to online CSAE material. Between January 2015 and March 2019, over 8.3 million unique first-generation images were added to the Child Abuse Image Database (CAID) \citep{nca2020}. From 2016 to 2017, the UK CSAE referrals bureau received, on average, over 4000 referrals per month, which is ten times more than in 2010 \citep{nca2017}. However, the true scale of CSAE is suspected to be much higher than recorded in official statistics \citep{nca2019}.

Some progress has been made in the auto detection and categorisation of CSAE material. These systems use hash algorithms to match seized images and videos with pre-recognised images/videos. Hashes of suspect images are compared with hashes stored in systems such as the Child Abuse Image Database (CAID\footnote{CAID is the UK's national child abuse image database that has been developed by the Home Office in cooperation with the police and industry partners \citep{kloess2019}.}), which facilitates the identification of previously known CSAE material \citep{kloess2019}. Not only do the obtained images have to be bit-identical, but images not matched with CAID have to be manually investigated and even where images match, a selection must be manually validated to ensure consistency.

The victim's age plays a crucial role in the prosecution of CSAE cases. Investigators must ascertain whether a subject -- the victim or suspect -- is a child or an adult, and quite often an assessment is made of the approximate age of the subject. The UK National Society for the Prevention of Cruelty to Children \citep{nspcc2020} and the UK Crown Prosecution Service \citep{cps2018} define a child as a person under the age of 18. The World Health Organisation (WHO) defines children as ``a person 10 years or younger" \ and adolescents as ``a person aged 10 to 19 years inclusive" \citep{who2013}. \cite{kloess2019} provide a more granular definition. They define the development stages of children as follows: early childhood (1-6), later childhood (6-10), younger adolescent (10-16) and older adolescent (14-17). \cite{kloess2019} definitions will be used throughout since they comply with the NSPCC and CPS definitions.

\subsection{Intelligent Forensics}

The term intelligent forensics, originally coined by \cite{irons2014}, refers to the ability of computers to learn particular forensic tasks from data using \textit{``intelligent"} methods, such as Machine Learning (ML), to enhance digital forensic processes. Notwithstanding a number of early studies acknowledging the value of AI in digital forensic investigations, there have been very few practical contributions and tools are still largely user-dependent. However, more importantly, very little research has been published on the optimisation of age estimation processes in DFIs.

\cite{peersman2016} present an image and video classification module for the identification of CSAE content using automatic filename categorisation, content classification (RGB-based skin detection) and hash-based detection. Their results suggest an accurate tool for the detection of CSAE material with false-positive rates of  7.9\% for images and 4.3\% for videos. This contribution may benefit from the utilisation of newer hashing algorithms, and reducing reliance on exact hash matching techniques given that cropping/cutting CSAE materials will render a different hash.

There appears to be a lot of scepticism for AI-based systems \citep{feldman2018, tasioulas2019}. \citeauthor{costantini2019} outlined that while forensic techniques, such as DNA analysis, are broadly accepted and trusted, AI-based support systems are still treated with mistrust and it may be difficult to convince the involved parties of the applicability of such support systems.

Unfortunately, these approaches do not provide a solution for the fundamental legal investigative problem: the estimation of a victim's age. Despite the importance of seizing and categorising CSAE material, the estimation of age and subsequently, the identification of victims is the central concern in law enforcement. Hence, intelligent forensic techniques for reducing the forensic backlog, categorising images, standardising information representation are, without a doubt, necessary and long overdue.

\subsection{Human Factors in Digital Forensic Investigations}

\paragraph{Cognitive Bias}

Forensic experts do not always produce consistent results and are sometimes prone to errors related to technological and human factors \citep{sunde2019}. Technological errors include faulty timestamps, data loss, and errors related to the lack of scientific data and research. Human factor related errors include cognitive bias, intentional bias and inadvertent errors.

The sources of cognitive bias are strong emotions, such as anger, frustration or confidence often due to the \textit{``underpinning of expertise"} \citep{dror2011}. Intentional bias and general errors are related to lack of motivation, training or incompetence. \cite{sunde2019} suggest that potential countermeasures can include psychological training, peer review or hypothesis-testing.

\paragraph{Psychological Stress}

There is a growing body of research which outlines the psychological impact of exposure to CSAE material. Investigators can develop traumatic stress caused by constantly processing and examining CSAE material \citep{sanchez2019}. \cite{powell2015} investigated the impact of observing CSAE material on 32 child exploitation investigators and found that this activity contributed to salient emotional, psychological, social and behavioural consequences. Similar findings have been reported in a study of 129 law enforcement officers by \cite{seigfried2018}. The results showed that contact with CSAE material cases leads to low self-esteem, impaired concentration compared to other uninvolved investigators, and/or Secondary Traumatic Stress (STS) which relates to symptoms which occur from observing or learning about a traumatic event \citep{bourke2014}.

\subsection{Age Estimation Techniques}

While face recognition has been extensively studied \citep{mathias2014, parkhi2015, rowley1998, turk1991, viola2004}, automatic and precise age estimation is still a comparably new research field. Recent deep learning age estimation techniques \citep{anand2017, antipov2016, chen2017, escalera2015, liu2017, rothe2015}, which have built upon early work on exact age estimation \citep{fu2010, guo2012, lanitis2002, lanitis2004}, have shown promising results.

Recently, there has been an increase in studies that leverage deep neural networks for face detection, classification of gender and age estimation from facial images \citep{gonzalez2018}. From simple examples, which use 3-layer neural networks \citep{agarwal2010}, to other more sophisticated ones, which use multiple neural networks \citep{dong2016} and/or networks that potentially consist of hundreds of layers \citep{huang2016}, have also been used successfully by numerous researchers.

Facial appearance is an essential trait in estimating human age \citep{geng2013}. Humans can usually accurately determine facial information such as identity, gender and race; however, the estimation of age is more challenging.

Recent progress in this field can be ascribed to the developments in the area of deep learning and the availability of large scale datasets \citep{parkhi2015}. Face recognition has benefited immensely from comprehensive and high-quality datasets, such as Labeled Faces in the Wild (LFW) and YouTube faces \citep{eidinger2014}. However, facial image datasets such as LFW cannot be considered for age estimation tasks due to missing age labels.

\subsubsection{Facial Image Datasets}

\begin{table*}[!t]
\caption {Comparison of different facial image datasets.}
\resizebox{\textwidth}{!}{%
\begin{tabular}{|l|l|l|l|l|l|l|l|l|l|}\hline
\textbf{Literary source} & \textbf{Dataset} & \textbf{Subjects (\female)} & \textbf{Images (\female)} & \textbf{Age} & \textbf{$\leq$ 16} & \textbf{$\leq$ 25} & \textbf{Max gap} & \textbf{P} & \textbf{A}\\ \hline
\cite{ricanek2006}&MORPH Album 1&515(95)&1724(294)&N/A&N/A&N/A&29&Y&Y \\ \hline
\cite{ricanek2006}&MORPH Album 2&13618(2159)&55608(8551)&16-77&N/A&N/A&5&Y&Y \\ \hline
\cite{lanitis2015}&FG-NET&82(34)&1002(395)&0-69&61.07&80.94&45&Y&Y \\ \hline
\cite{chen2014}&CACD&2000&163446&16-62&N/A&N/A&10&Y&Y \\ \hline
\cite{liu2016}&CAFE&901&4659&N/A&N/A&58.04&N/A&Y&Y \\ \hline
\cite{bianco2017}&LAP&1010&3828&N/A&N/A&N/A&80&Y&Y \\ \hline
\cite{rothe2015}&IMDB-WIKI&22741&524230&N/A&N/A&N/A&N/A&Y&Y \\ \hline
\cite{escalera2015}&LAP&N/A&4699&N/A&N/A&N/A&N/A&Y&N \\ \hline
\cite{deb2018}&CLF&919(315)&3682&2-18&N/A&100&6&N&N \\ \hline \cite{grd2016}&ageCFBP&287(151)&1655&0-25&~40&100&N/A&N&Y \\ \hline
\cite{founds2011}&MEDS&518(52)&1309&15-69&N/A&34.37&N/A&N&Y \\ \hline
\cite{eidinger2014}&OUI-Adience&2284&26580(12826)&0-60&N/A&38,67&N/A&Y&Y \\ \hline
\cite{fu2008}&UIUC-IFP-Y&1600(800)&8000(4000)&0-93&N/A&N/A&N/A&N&Y \\ \hline
\cite{gupta2012}&FSAR&N/A&432&N/A&N/A&N/A&N/A&Y&N \\ \hline
\cite{choi2011}&BERC&390&390&3-83&N/A&N/A&N/A&N&Y \\ \hline
\cite{somanath2011}&VADANA&43(17)&2298&0-78&N/A&N/A&N/A&Y&Y \\ \hline

\multicolumn{8}{c}{\female $\,\to\,$ female; P  $\,\to\,$ Publicly available. A  $\,\to\,$ Applicable to age estimation research}
\end{tabular}  }
\end{table*}

One of the most pressing issues in facial age estimation is the lack of sufficient training data for many different age groups \citep{rothe2016}. Because age estimation methods commonly depend on a process of learning and training, the selection of the right training data is a crucial step that defines the subsequent performance of a classifier \citep{gonzalez2018}. In the case of the age estimation of CSAE victims, a well-trained neural network can still be inaccurate if trained with the wrong age groups.

One of the oldest and largest databases are the MORPH albums which consist of 1724 and 55608 images, respectively, and can be mainly used for age-gap verification (Figure 1). MORPH Album 2 has a small age-gap issue compared to Album 1, which could pose a problem for the age gap verification accuracy.

The Face and Gesture Recognition Network (FG-NET) database was published in 2004 and is a valuable and important contribution to the field. The dataset contains 1002 images from 82 different individuals with ages ranging between 0 to 69. Approximately 61\% of the images show subjects that are younger than 16 and about 81\% are younger than 25. The publisher's aim, was to contribute to research technology development in the field of face recognition by providing a publicly available dataset of facial images. However, one limitation of the FG-NET database is the lack of racial diversity and the relatively small size.

\cite{rothe2015} proposed a new method called Deep Expectation (DEX) and crawled over 500,000 celebrity images from both Wikipedia and IMDB to produce a database of images annotated with age and gender labels. \cite{rothe2015, rothe2016} use a model with 101 output neurons, including age ranges of 0 to 100. The IMDB-WIKI dataset benefits from containing a high number of images. Unfortunately, the dataset suffers from a lack of training images (training density) for low age ranges, i.e. it is inaccurate and therefore unhelpful in facial age estimation of very young children. However, \cite{rothe2015} achieve low MAEs in estimating the ages of mid-to-late teenagers, which is particularly important when attempting to distinguish jurisdictional boundaries, for example, distinguishing CSAE material from adult pornography. Furthermore, the authors do not explicitly employ facial landmarks, which, as they state, could improve the performance of DEX. Facial landmarks are fiducial facial key points that are used to analyse facial images. Facial landmark detection algorithms automatically identify the locations of eyes, nose, mouth, wrinkles and other unique features \citep{wu2019facial}. Yet, even without relying on landmarks and by handling small occlusions robustly, DEX significantly outperforms the human reference on facial age estimation of adults.

Table 1 shows a detailed list of facial image datasets for two different purposes – age estimation and age-gap verification. Figure 1 shows the purpose of each dataset and the different factors of diversity. Although racial diversity would benefit the overall accuracies of age estimation, it cannot be considered as a diversity factor due to the limited information about race distributions in publications in this domain. Hence, the diversity factor is calculated by considering the gender and age distributions.

\cite{grd2016} acknowledged that the accuracy of child age estimation suffers from lowered accuracies due to the lack of training data for these specific age groups. The majority of available datasets do not offer images of individuals from birth to adulthood.

\cite{grd2016} chose to manually collect personal data of minors, instead of crawling publicly available data from the internet. This method requires parental consent, involves fundamental legal considerations and is not scalable, whereas related studies such as the research from \cite{antipov2016} and \cite{liu2016} apply automated data crawling methods to enhance the efficiency of the image collection process. The datasets of both \cite{antipov2016} and \cite{grd2016} have remained private.

The previous studies reveal that the lack of facial images of children poses a big problem for the accuracy of age estimation techniques, which could be applied to DFIs. As \cite{huang2008} conclude, training and testing face recognition and age estimation algorithms on highly diverse sets of faces is crucial to achieving the best possible results. Hence, the value of studying the above areas of literature will support this research on developing perfectly tailored datasets for juvenile age estimation.

\begin{figure*}[!ht]
\begin{center}
\includegraphics[scale=0.40]{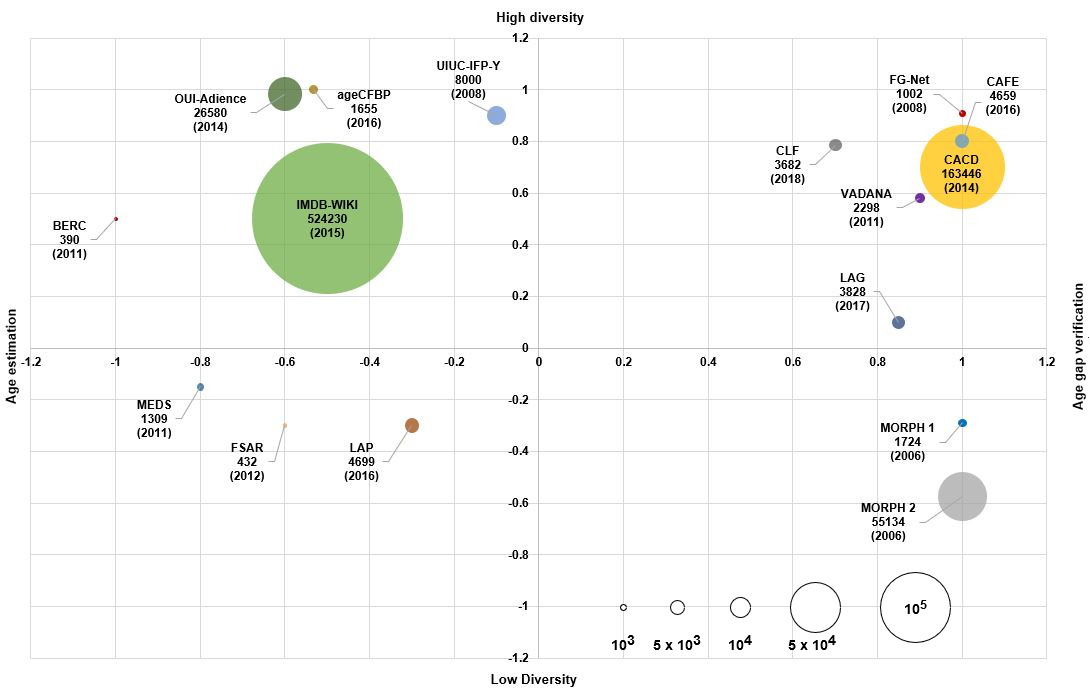}
\caption {Bubble diagram of facial image datasets showing the name, size, year, diversity and purpose.}
\end{center}

\end{figure*}\label{fig:dataset_bubbles}

\subsubsection{Age Estimation Using Neural Networks}

The estimation of a person's age can be achieved using many different methods. This section focuses specifically on neural networks (NNs).
A neural network is a neuron-connectivity pattern inspired by the neural structure of mammals \citep{zhang2019}. 
Neural networks firstly proved their ability in object recognition in the ImageNet competition \citep{krizhevsky2012}. They can be used for many different tasks that involve learning from images, such as medical image analysis, exchange rate predictions but also age estimation. From a high-level perspective, age estimation methods can be divided into machine learning approaches based on feature sets and deep learning pattern recognition techniques \citep{anand2017}.

\paragraph{Classification and Regression}

Age estimation can be seen as a classification or regression problem \citep{chen2017}. In an age classification task, a facial image is assigned to a particular group by a multi-class classification algorithm. A regression task outputs a discrete age estimation value \citep{dong2016}.

\cite{rothe2015} argue that age estimation is a piece-wise regression problem, since age is a continuous variable, although they still train convolutional neural networks (CNNs) for classification into 101 year labels [0...100]. When training for classification, it is vital that each age range covers the same number of years and is compromised of the same number of training images \citep{rothe2016}.

Classification models include Support Vector Machines (SVM), CNNs, DNNs, Nearest Neighbours, whereas common regression methods include Support Vector Regression (SVR) and Quadratic Regression (QR).

\paragraph{Convolutional Neural Networks}

Deep CNNs have shown promising results for computer vision tasks on images \citep{park2016}. Recently, many studies have employed CNNs to predict the age of an individual from a single facial image \citep{anand2017, antipov2016, chen2017, feng2019, rothe2015}. Their methodology is generally identical; by combining a series of convolutional, pooling and fully or partially connected layers as well as activation functions, the final output value is calculated (see Figure \ref{fig:network_1}).

CNNs apply, as the name suggests, discrete convolution (1) to extract a set of local features from a facial image. A convolutional layer performs the mathematical convolution over the matrices of preceding layers in order to extract facial features from the image. Facial features can be divided into high-level, medium-level and low-level features which respectively describe the facial structure, more detailed features (eyes, ears, nose, mouth) and very detailed features (edges, dark spots, wrinkles) \citep{liu2015}.

\begin{equation}
(f\ast g)(t)=\sum_{\tau=-\infty}^{\infty} f(\tau)g(t-\tau)
\end{equation}
\label{DiscreteConvolution}

Equation (1) describes the sum of the mathematical products of two (shifted) functions $f(t)$ and $g(t)$ for every point $\tau$ in time. Convolution in digital image processing is performed with a filter array (filter matrix). This filter array and every shifted version of the array are multiplied with the original image. The result is a matrix that describes an extracted feature.

The pooling layers are responsible for down-sampling the image and thus reducing the complexity. This process can be viewed as decreasing the resolution of an image. Max-pooling is one method that is commonly used within CNNs and refers to summarising the distinct regions of an image to a single value. Generally, a 2 $\times$ 2 pixel region is combined and the highest pixel value is forwarded to the next layer \citep{albawi2017}. \cite{krizhevsky2012} were using an overlapping pooling scheme for their famous ImageNet classification. This method, compared to non-overlapping pooling, can reduce error rates by overlapping the grids with one another. Another advantage observed by the authors is that overlapping pooling layers are slightly reducing the probability of overfitting, which CNNs are generally prone to.

A fully-connected layer, as the name suggests, directly connects each neuron in a layer to every single neuron in another layer. In order to redirect the matrix output of the convolutional and pooling layers into a dense layer, the layer needs to be flattened first.  They are commonly placed at the rear end of a CNN to produce the final classification decision \citep{dung2019}. A disadvantage of fully-connected layers is the high computational complexity compared to partially-connected layers. Hence, many modern networks are limiting the number of fully-connected layers to a minimum (including the VGG-16 CNN used in this research).

Every convolutional or fully-connected layer usually feeds the output into the Rectified  Linear Unit (ReLU) non-linearity function \citep{krizhevsky2012}.
To achieve non-linearity and therefore adjust the output values, different mathematical functions such as tanh, sigmoid, or ReLU can be applied. The latter is the most commonly used in today's CNNs since it is more computationally efficient than other functions. The ReLU ensures that all values that are $ \leq 0$ are becoming 0 and the remaining values stay in a 1:1 proportion.

When training a network, each iteration optimises the network by back-propagating the learned information. The process of forward-propagation is merely feeding the input images into the network and calculating a variable in the output layer. However, the weights in a network need to be adjusted accordingly based on the accuracy of the output layer. Hence, the back-propagation of the errors is necessary. The error value is determined by calculating the difference between the expected value and the actual output value \citep{shen2016}.

\subsection{Summary}

The background outlined herein highlights the need for the creation of more tailored datasets for digital forensic investigations. Several datasets could be more applicable to DFIs if the creators paid close attention to image diversity, especially as this factor is crucial for training deep neural networks.

Further research is required to gain a better understanding of the most effective facial image dataset composition for achieving more accurate age estimation results on CSAE images. The next stage of this research will empirically evaluate the accuracy of an existing algorithm and determine the efficacy of the prototype.

\section{Methodology}\label{Methodology}

An age estimation prototype was developed and evaluated using statistical metrics to assess the accuracy of the prototype and compare the results to existing age estimation solutions. Additionally, we collected and analysed qualitative responses to determine the efficacy and usefulness of the approach to digital forensic practitioners. This was intended to help identify further improvements to the age estimation prototype.

\subsection{Framework for the Development of a Prototype}

The availability of adequate datasets is a key challenge in this research area. There are a number of strictures which prevent the gathering of such data. Children in the UK can give their own consent under the GDPR at the age of 13. Any child younger than that relies on parental consent \citep{ico2020}.

The background section revealed that there are existing datasets that can be considered for this research. These datasets can be used to both serve as input data for training new models or fine-tuning pre-trained models. The large age estimation dataset ``IMDB-WIKI" was chosen to serve as a basis of the age estimation process. 

We adopted the approach demonstrated by \cite{lanitis2004} and \cite{liu2016} and merged the  dataset from \cite{lanitis2004} and the newly collected dataset. This method has been selected because (a) no legal issues arise when collecting public images\footnote{As long as the dataset is not made public. If we decide to publish the collected dataset, legal factors such as copyright must be taken into account.} and (b) several existing datasets have pertinent quality and quantity that fulfils the requirements of this research.

However, some challenges arise with this approach, such as potential redundancies or missing age labels. Redundant images can hardly be identified because the image hashes are almost always different for the newly collected dataset (due to crop sizes). Furthermore, missing age labels cannot be reconstructed; hence some images might need to be discarded.

The newly collected dataset contained some redundancies that were removed based on the filename structure \textit{firstname\_lastname\_age.filetype} to end up with one image per individual. The same approach has been applied to the 1002-image FG-NET dataset.

\subsection{Testing the Prototype Accuracy}

\subsubsection{Statistical Metrics}

In order to quantitatively analyse the accuracy of the prototype, we applied selected statistical methods.

Perhaps the most popular metric for evaluating systems of age estimation is the Mean Absolute Error (MAE), which is defined as a ``mean value of absolute differences between predicted ages x-hat and real (biological) ages" \ \citep{antipov2016}. The MAE (2)  is the industry-standard measure for age prediction errors \citep{anand2017, antipov2016, chen2017, rothe2015, rothe2016}. The measurement can be described as follows:

\begin{equation}
MAE = \dfrac{1}{N} \sum_{i=1}^{N} |\hat{x_i} - x|
\end{equation}
\label{MeanAbsoluteError}

In addition to the MAE, many researchers also adopt the Cumulative Score (CS) as a performance measure \citep{chen2017, han2013, weng2013}. The CS (3) indicates the percentage of images for which the error is less than a certain number of years \citep{han2013}. The higher the CS, the more accurate is the age estimation system. The CS measurement can be implemented by either defining an absolute age range (e.g. \textpm 2 years) or a neighbouring age range (e.g. the two neighbouring age classes). This measurement can be described as follows:

\begin{equation}
CS(l) = \dfrac{M_{e\leq l}}{M}
\end{equation}
\label{CumulativeScore}

Another commonly used metric is the $\epsilon$-error (4). This quantitative measure can be used for datasets where there is no ground truth age, but instead, a group of people vote for the age label. The error measures the standard deviation $\sigma$ of the guessed age. The final $\epsilon$-error represents the average over all images, 0 being a perfect prediction and 1 being a wrong prediction \citep{rothe2016}. This measurement can be described as follows:

\begin{equation}
\epsilon = 1 - e^{-\frac{(x-\mu)^2}{2\sigma^2}}
\end{equation}
\label{EpsilonError}

MAE and CS were applied in this research. The $\epsilon$-error was not used because no apparent age estimation has been carried out. However, this metric could be appropriate for future research in this area.

The MAE was used to evaluate the absolute accuracy of the prototype. A small MAE means that the estimated age is close to the real age and thus a reasonable estimate. On the other hand, a high MAE value indicates inaccurate results. The CS was used to evaluate the overall performance of the age estimation prototype within a particular age interval. The higher the CS measurement, the more accurate is the age estimation system.

In general, the validity and reliability of accuracy evaluation methods can be ensured by correlating the different statistical methods and thus proving the trustworthiness of the results. By combining the MAE and the CS, a high level of quality concerning the consistency of the results can be achieved.

\subsection{Testing the Prototype Efficacy}

The researchers were mindful that this was one of the first contributions in this area and that it was important to determine the value this might add to an investigation, thereby motivating further development in this field. To this effect, qualitative interviews were conducted with two investigators from West Midlands Police and two academic researchers in order to aid our understanding of the efficacy of our approach.

\begin{table}	
\caption {Interview participant information.}
\scriptsize
\begin{center}
\begin{tabular}{|C{5mm}|C{20mm}|C{25mm}|C{12mm}|}\hline
    \textbf{Ref} & \textbf{Role} & \textbf{Department} & \textbf{Years of Experience}\\ \hline
    A1&Higher Forensic Analyst&Digital Forensic Unit&14\\ \hline
    A2&Higher Digital Forensic Officer&Digital Forensic Unit&4\\ \hline
    B&Director of Research and Development&Forensic Research&10\\ \hline
    C&Principal Lecturer&School of Law/Policing/Social Sciences&15\\ \hline
\end{tabular} 
\end{center}

\end{table}\label{tab:interview_participants}

The interview transcripts were transcribed using manual intelligent verbatim transcription and narrative content analysis was performed subsequently. Narrative content analysis is commonly used to underline common themes between different participants and is necessary to get initial insights into the qualitative data.




\subsection{Limitations and Potential Problems}

As with the majority of studies, this research is also subject to limitations and potential problems that arise from both limited access to data and time constraints. These constraints relate to racial diversity in datasets and reliability issues.

The failure to test the prototype on a more diverse dataset could lead to racial bias due to unequal distribution of images. Future research could mitigate this problem by collecting an image dataset with a strong focus on the equal distribution of factors such as origin and skin colour of the subjects.

With more resources and time, the data collection process may be further optimised by developing an automated dataset generator that selects the most useful images from various existing datasets based on parameters such as age, race and image quality. This could benefit the study with regard to a more precise data distribution and a better overview of the correlation between a particular set of input images and the outcome, i.e. the estimation accuracy.

\begin{figure*}[!ht]
  \centering
  \begin{minipage}{0.48\textwidth}
    \includegraphics[width=0.8\textwidth]{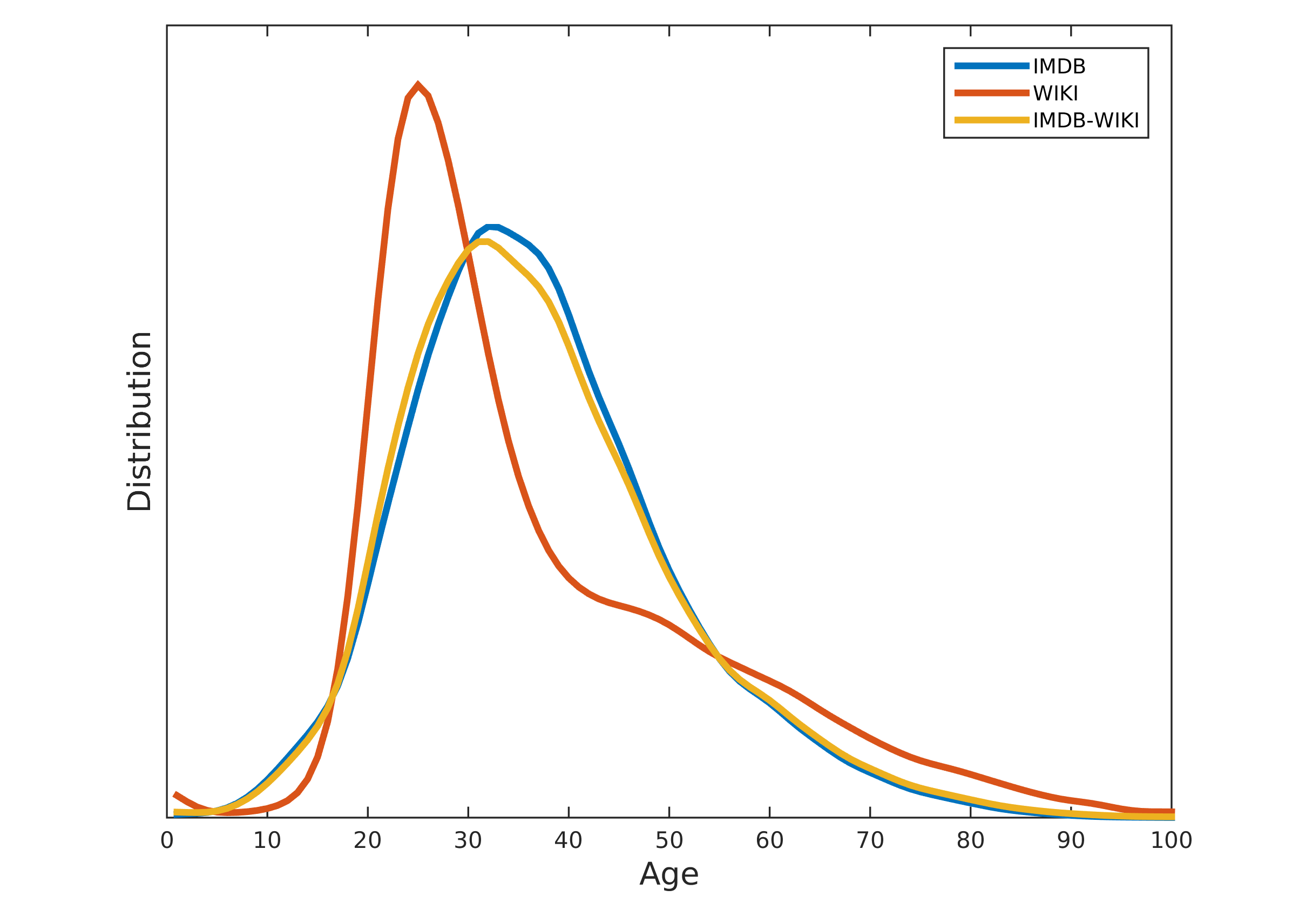}
    \caption{IMDB-WIKI age distribution function \citep{rothe2016}.} \label{fig:imdb_distribution}
  \end{minipage}
  \hfill
  \begin{minipage}{0.45\textwidth}
    \includegraphics[width=0.8\textwidth]{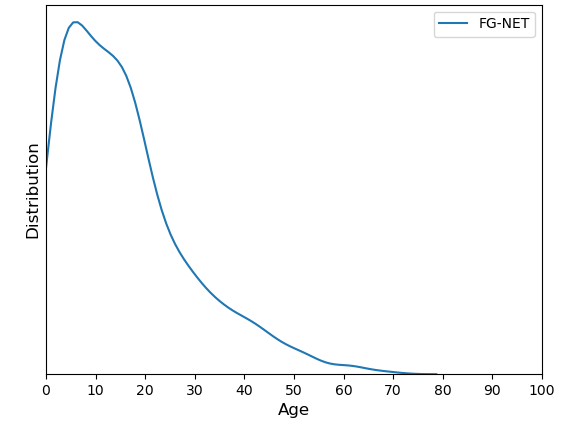}
    \caption{FG-NET age distribution function.} \label{fig:fgnet_distribution}
  \end{minipage}
\end{figure*}

The image preprocessing stage might be subject to reliability issues. Due to time constraints of the study, we cannot accurately determine the perfect crop size and rotation angle for every image to achieve the best possible estimation results. Thus, the CNN accuracy might deviate slightly from the best possible accuracy. To mitigate this issue, future research could incorporate a more sophisticated image preprocessing tool.

\section{Experiment Design}\label{Experiment Design}

The experiment tests the preliminary stage of an age estimation prototype that can be deployed in DFIs. Furthermore, the experiment includes a performance analysis of a pre-trained CNN (with the IMDB-WIKI dataset) and attempts to show how this CNN is performing on the newly collected dataset.

The main features of the prototype can be described as follows:

\begin{enumerate}
\item Take an input image in any image format (.jpg, .png, .bmp, ...).
\item Classify the input image into an age category pre-defined by the CNN (0-100 years, one for each year).
\item Print out a single number age estimation result within the range 0-100.
\end{enumerate}

\begin{table}[!b]
\centering
\footnotesize
\caption {Required Python packages.}
\begin{tabular}{|p{25mm}|p{11mm}|p{80mm}|}\hline
\textbf{Package} & \textbf{Version} & \textbf{Description} \\ \hline
Pillow&7.0.0&Python Imaging Library fork.\\ \hline
PyWavelets&1.1.1&Open source wavelet transform software.\\ \hline
cycler&0.10.0&Composable style cycles.\\ \hline
decorator&4.4.2&Signature-preserving function decorators.\\ \hline
imageio&2.8.0&Read and write image data.\\ \hline
kiwisolver&1.1.0&Cassowary constraint solving algorithm.\\ \hline
matplotlib&3.2.1&Creating static, animated, interactive visualisations.\\ \hline
networkx&2.4&Create and manipulate complex networks.\\ \hline
numpy&1.18.2&Package for scientific computing.\\ \hline
pip&19.0.3&Package installer for Python.\\ \hline
protobuf&3.11.3&Package for Google’s data interchange format\\ \hline
pyparsing&2.4.6&Create and execute simple grammars.\\ \hline
python-dateutil&2.8.1&Extends the standard datetime module.\\ \hline
scikit-image&0.16.2&Collection of algorithms for image processing.\\ \hline
scipy&1.4.1&Software for mathematics, science, and engineering\\ \hline
setuptools&40.8.0&Package for facilitating packaging Python projects.\\ \hline
six&1.14.0&Helps smoothing differences between Python versions.\\ \hline
\end{tabular} 
\label{tab:libraries}
\end{table}

\subsection{Choice of Technology}

The Debian 10.3 Linux distribution was used to set up the development environment. The Caffe deep learning framework has several dependencies (Table \ref{tab:libraries}). In addition, multiple libraries are needed to process input images, present the data and build a GUI, the \cite{rothe2015} pre-trained Caffe model for real age estimation was used. The model ``contains" \ weights adjusted to over 500,000 training images. Hence, using this Caffe model is a considerable time-saving.

\subsubsection{Datasets}

\begin{figure}[!b]
\centering
\caption {Preliminary graphical user interface of the age estimation prototype.}
\includegraphics[scale=0.27]{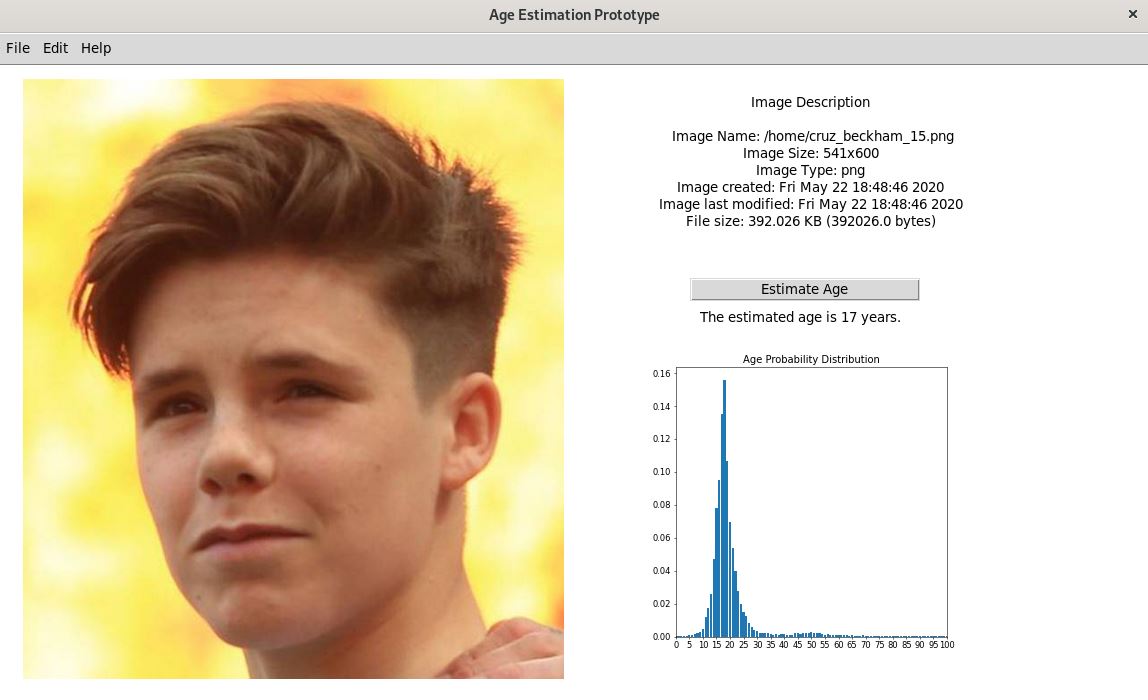}
\end{figure}\label{fig:age_estimation_gui}

The characteristics of the IMDB-WIKI dataset have already been laid out in detail in the Background section. However, an important detail that helps with the interpretation of the results in the subsequent chapter is the distribution of age within the IMDB-WIKI dataset as shown in Figure \ref{fig:imdb_distribution}. This plot gives insights into the age density of images. The testing data consists of the newly collected dataset of 245 images and an existing dataset of 82 images.
Images from the FG-NET dataset (fourth highest diversity score) have been merged with the new dataset in order to achieve a high-quality testing set (Figure \ref{fig:fgnet_distribution}). Since the FG-NET dataset contains images from 82 unique individuals, one image per person has been randomly selected where the person depicted is 20 years old or younger.

\begin{figure*}
    \begin{center}
    \caption {IMDB-WIKI VGG-16 Convolutional Neural Network architecture (split into six parts).}
    \includegraphics[scale=0.33]{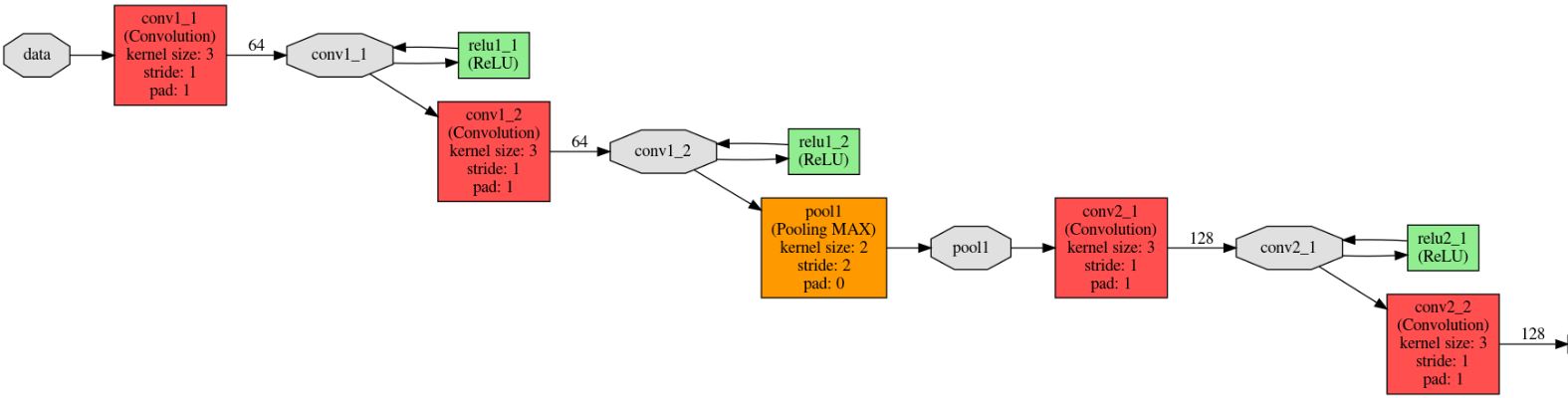}
    \label{fig:network_1}
    \end{center}
    
    \begin{center}
    \includegraphics[scale=0.27]{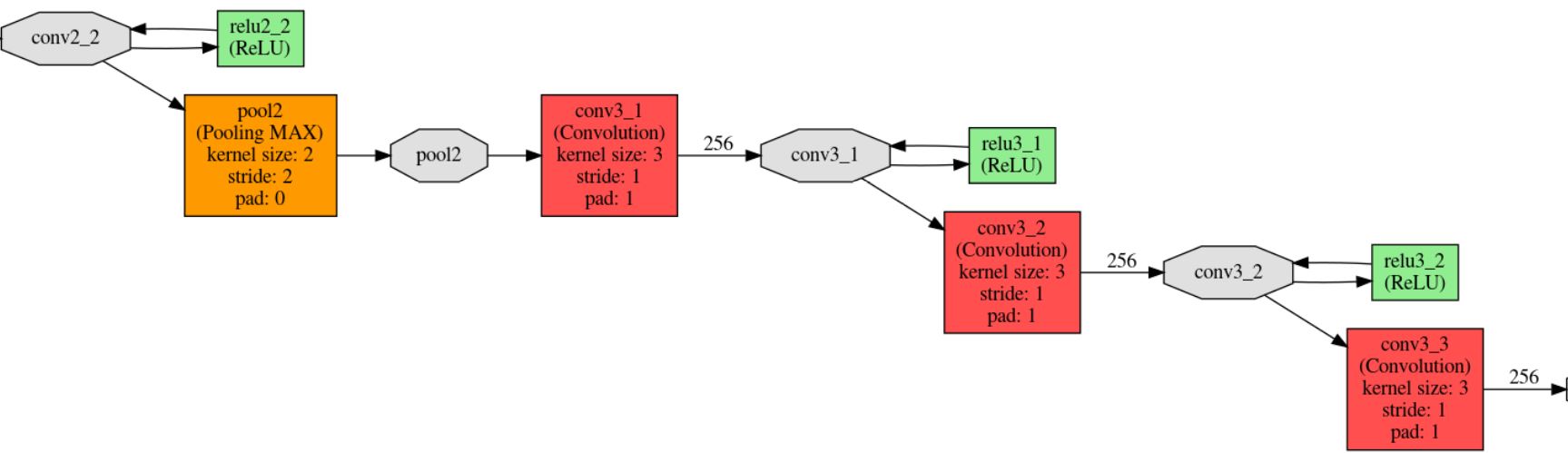}
    \label{fig:network_2}
    \end{center}
    
    \begin{center}
    \includegraphics[scale=0.32]{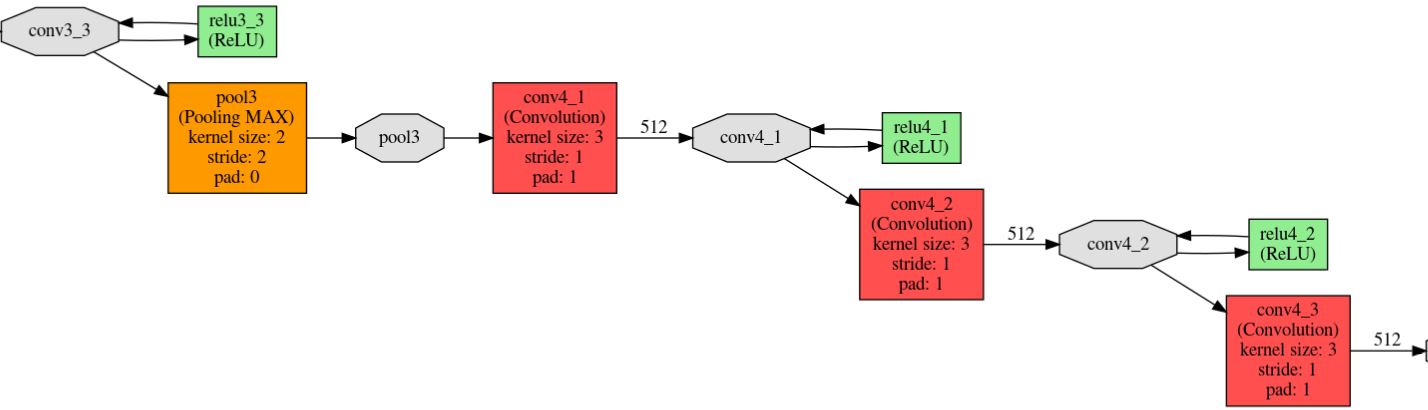}
    \label{fig:network_3}
    \end{center}
    
    \begin{center}
    \includegraphics[scale=0.32]{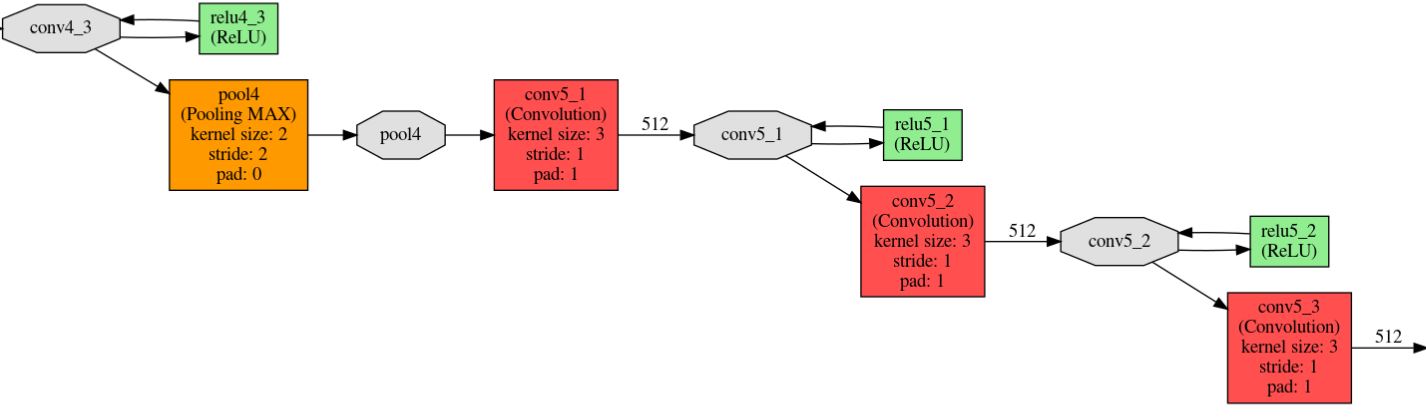}
    \label{fig:network_4}
    \end{center}
    
    \begin{center}
    \includegraphics[scale=0.32]{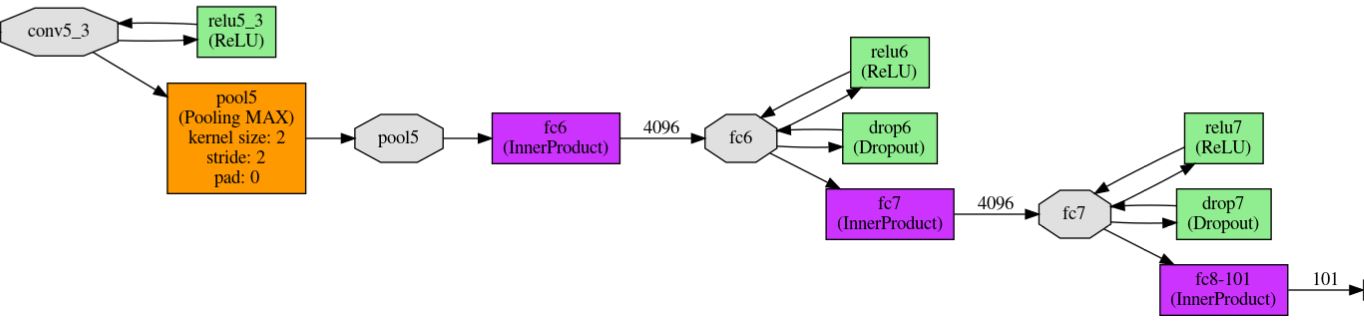}
    \label{fig:network_5}
    \end{center}
    
    \includegraphics[scale=0.25]{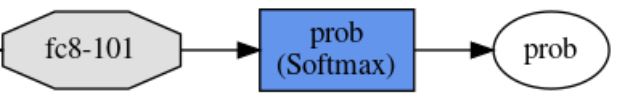}
    \label{fig:network_6}
    
\end{figure*}

\subsubsection{Data Processing}

\begin{figure*}
\begin{center}
    \caption {High-level software component diagram of the deep learning age estimation prototype.}
    \includegraphics[scale=0.6]{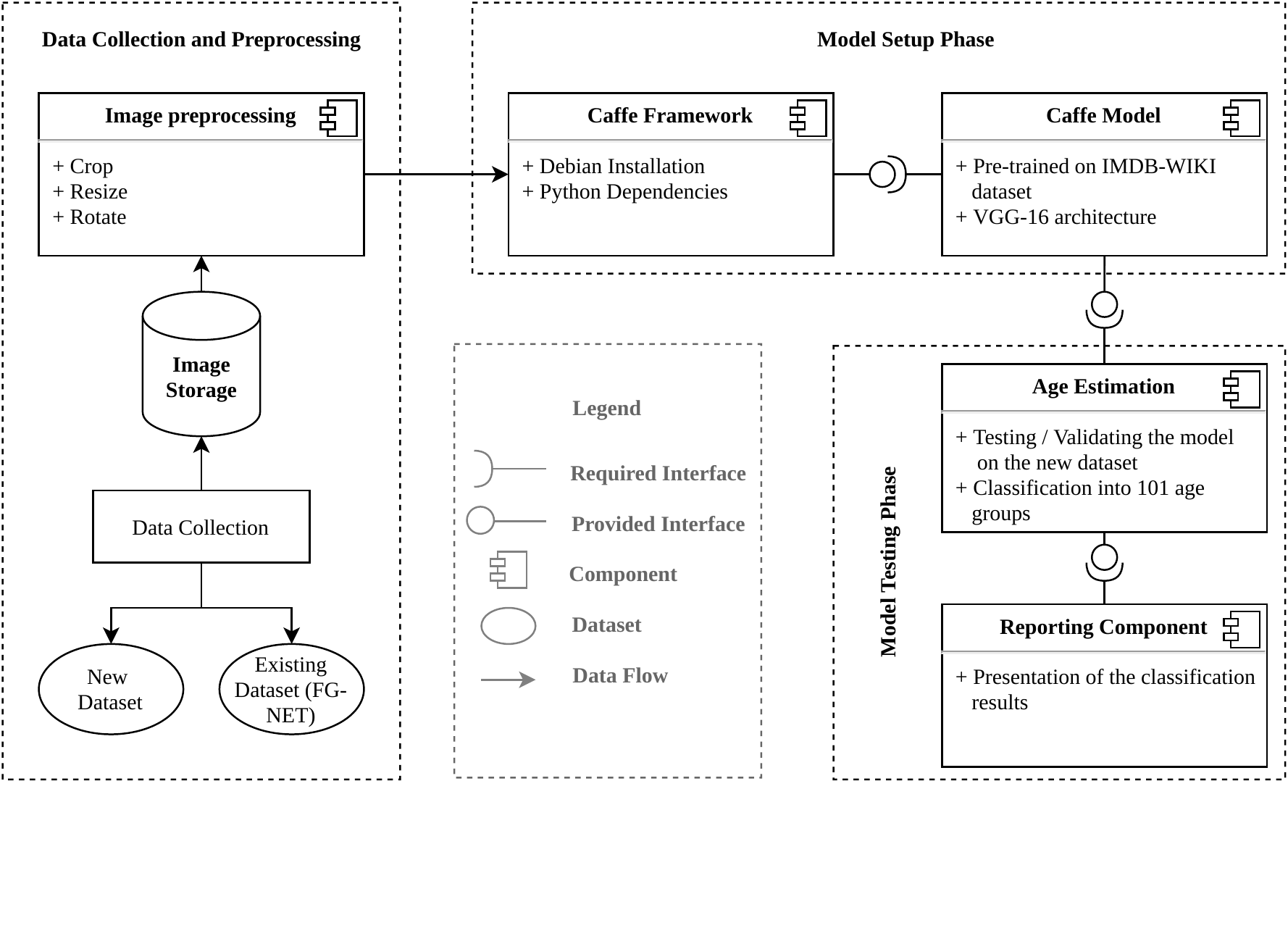}
    \label{fig:prototype_component_diagram}
\end{center}
\end{figure*}

The structure of the deep learning framework is presented as a high-level software component diagram (Figure \ref{fig:prototype_component_diagram}). Each component can be interpreted as a higher abstraction than a class. The VGG-16 architecture and the built-in 16 convolutional weight layers, is a convolutional network for large-scale image and video recognition. The Caffe deep learning framework was developed by \cite{jia2014} with the Berkeley AI Research team and is one of the most widely used frameworks for image classification.

The proposed model architecture, consisting of the pre-trained model and testing dataset, has two main advantages over traditional training methods: (a) simplicity and low computational cost, because Caffe is among the fastest CNN implementations available \citep{jia2014} and (b) the likelihood of overfitting in future training processes is reduced by adopting a well-balanced pre-trained model.
The main reason why the Caffe framework has been selected over other platforms such as Tensorflow or PyTorch is that the pre-trained model is readily available as a Caffe model. Furthermore, Caffe provides the option of simply converting the model into Tensorflow.
The image preprocessing has been conducted using the \cite{rothe2015} face extraction Matlab script. However, since many images were not cropped accordingly, the image preprocessing stage also involves manual crops and resizes.

\subsubsection{Network Architecture}

This experiment uses a CNN with a VGG-16 architecture \citep{rothe2016}. It has a deeper architecture than its predecessors with 16 layers in total (13 convolutional and three fully-connected)\footnote{A more detailed description of the different CNN components can be found in the Background section.}. A unique characteristic of this network is its relatively small convolutional filters with a size of 3$\times$3 pixels. This enables the net to capture more superficial geometrical structures rather than more complex ones. Lastly, the expected value is computed by applying the Softmax normalisation function to the output probabilities of the 101 neurons (one for each age class).

The network architecture is presented in Figure \ref{fig:network_1}. The network was created by running the \textit{\textbf{draw\_net.py}} script that is integrated into the Caffe framework: \newline

\noindent
\begin{verbatim}
python3 CAFFE\_DIRECTORY/python/draw_net.py \
MODEL\_DIRECTORY/age.prototxt \
OUTPUT\_DIRECTORY/network.png
\end{verbatim}

\subsection{Design and Development}

The Caffe model and \textit{prototxt} file were imported using the Python function \textit{estimate\_age()}. This function represents the core logic of the estimation prototype and includes loading the image, image processing and computing the output values.

The prototype does not only provide an estimated age integer value but also a probability distribution function, which was plotted using \textit{matplotlib}. This graph, shown in Figure \ref{fig:age_distribution_example}, represents each age class (0-100) on the x-axis and the associated probability value (0-1) on the y-axis.

\section{Results and Analysis}\label{Results and Analysis}

This chapter outlines the quantitative results from the evaluation of the pre-trained Caffe model on the new testing set. First, the age estimation results of individuals aged 0 to 20 are presented. Secondly, the corresponding MAEs and cumulative scores are reported. Lastly, the accuracy of the prototype is evaluated and analysed.

\subsection{Evaluation of Results}

The results are displayed as single-number age estimation values that represent the output of the age estimation prototype.

From a high-level perspective, results are varying across different age classes and from image to image. The obtained results underline the fact that the pre-trained model has a low image density in the lower age ranges (0-16). This model characteristic is responsible for most of the inaccurate classifications. The notably large errors can also partly be traced back to insufficient image quality and facial rotations that cannot be eliminated during the image preprocessing phase. A part of the dataset was purposely filled with imperfect images to adequately reflect genuine DFIs. In the area of age estimation, imperfect images include facial rotations, partial occlusions, zoomed faces, blurry images and any other characteristic of an image that can reduce the accuracy of age estimation software \citep{elmahmudi2019}.

\begin{figure}
\begin{center}
\caption {Example of face images in the new testing dataset.}
\includegraphics[width=0.7\columnwidth]{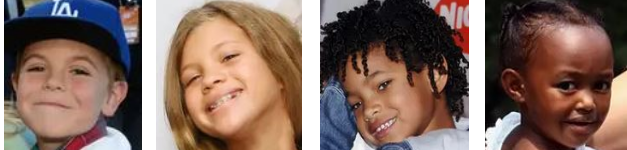}
\end{center}
\end{figure}\label{fig:example_images}

Another striking trend can be observed that the majority of estimations is higher than the real age, which can be related to the high image density in the IMDB-WIKI dataset in the [30; 40] age interval. The so-called estimation-shift is a common phenomenon amongst deep learning models that lack training images in a certain range. This trend is especially prevalent amongst the children in the age class (0-10) and younger adolescents (10-16). However, the lower age ranges are not particularly relevant for DFIs because investigators are generally able to estimate the age of children correctly.

\begin{figure}
\begin{center}
\caption {Age distribution example.}
\includegraphics[width=0.5\columnwidth]{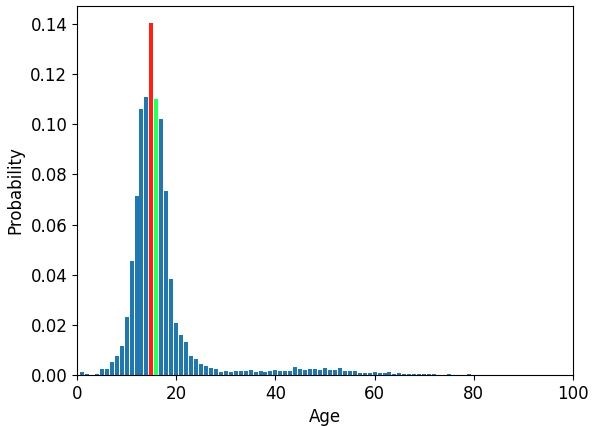}
\label{fig:age_distribution_example}
\end{center}
\end{figure}

It can also be observed that images that are classified with a small error of $\pm$ 2 years are often within the second or third ``guess" \ of the neural network. So, for example, when a 16-year-old is estimated to be 15 years old (Figure \ref{fig:age_distribution_example}), it is common that the real age class is within the three highest probabilities in the age probability distribution function. The red bar represents the wrongly calculated age of 15, the green bar shows the real age.

The findings also show that there is a correlation between the estimation accuracy and the distribution of the age probability curve. Results that are relatively inaccurate, such as the age estimation of very young children, often result in a flat distribution, i.e. a probability curve that is evenly distributed. This outcome reflects the ``indecisiveness" \ of the CNN. On the other hand, accurate results usually produce a steep age distribution curve, as shown in Figure \ref{fig:age_distribution_example}.

\subsection{Performance Analysis}

As Figure \ref{fig:mae_figure} depicts, the MAE between age ranges differs greatly. Estimations in the lower age ranges, in particular, are significantly deviating from the ideal estimation line. One can suggest that the low image density of the pre-trained model in the age range [0; 20] is  likely a contributing factor to the high MAEs shown in Table \ref{tab:mae_cs_table} and Figure \ref{fig:mae_figure}. The proportion of images in the age range [0; 7] that are correctly classified is 0\%. This indicates that images in this age range are scarce in the training dataset and thus the neural network does not know how to cope with this kind of input data. The MAE peaked at x = 0 years old with a value of 14.00 and x = 17 accounts for the minimum MAE of 1.79. Although the plot shows MAEs ranging into the sub-zero range, estimations smaller than 0 are not possible. This is merely a way of presenting the absolute value of the MAE.

These results are particularly advantageous since digital forensic investigators struggle the most with images depicting younger and older adolescents (10-17), based on the interviews with digital forensic investigators. In many cases, investigators are able to correctly estimate the age of children, whereas the age estimation prototype makes mostly inaccurate decisions in this range. The MAE is decreasing around the age of consent boundary and thus the estimation accuracy is increasing in this critical range for investigators. \\

\begin{figure*}
\centering
\begin{center}
\caption {Estimated vs real age of exemplar images.}
\includegraphics[width=0.7\columnwidth]{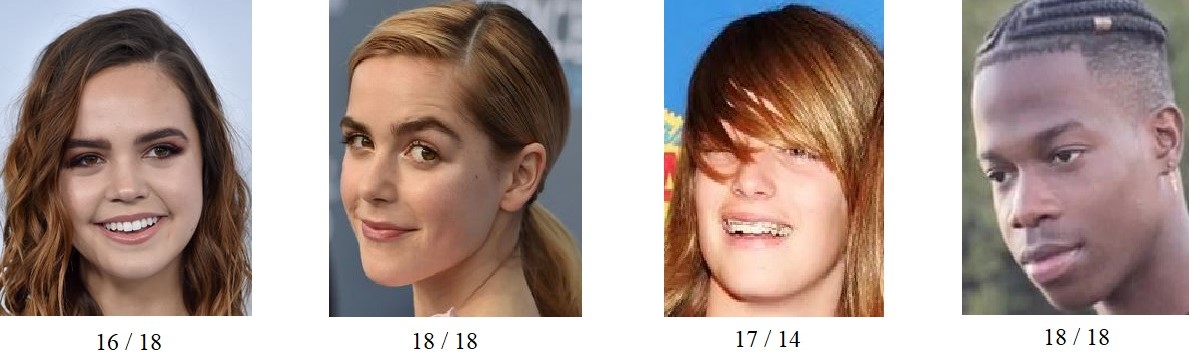}
\end{center}
\label{fig:example_estimations_1}
\end{figure*}

\begin{table}[h]	
\caption {Mean Absolute Errors and Cumulative Scores per age class.}
\scriptsize
\begin{center}
\begin{tabular}{|C{9mm}|C{10mm}|C{12mm}|C{12mm}|C{12mm}|}\hline
\textbf{Age Class} & \textbf{MAE} & \textbf{CS=1} & \textbf{CS=2} & \textbf{CS=3}\\ \hline
0&14.0&0&0&0\\ \hline
1&8.73&0&0&0\\ \hline
2&8.68&0&0&0\\ \hline
3&8.42&0&0&0\\ \hline
4&6.94&0&0&0\\ \hline
5&7.0&0&0&0\\ \hline
6&5.17&0&0&0\\ \hline
7&3.58&0&25&66.66\\ \hline
8&3.15&15.38&53.85&53.85\\ \hline
9&5.0&23.08&23.08&30.77\\ \hline
10&3.17&25&41.66&75\\ \hline
11&2.8&26.66&73.33&73.33\\ \hline
12&1.86&64.28&71.43&85.71\\ \hline
13&1.92&53.85&69.23&69.23\\ \hline
14&3.12&47.06&58.82&85.71\\ \hline
15&2.72&16.66&38.88&77.77\\ \hline
16&2.45&65&70&80\\ \hline
17&1.79&52.63&84.21&89.47\\ \hline
18&2.32&45.45&63.64&81.82\\ \hline
19&3.22&22.23&38.89&50\\ \hline
20&2.71&33.33&52.38&57.14\\ \hline
\end{tabular}
\end{center}
\label{tab:mae_cs_table}
\normalsize
\end{table}

\begin{figure*}
  \centering
  \begin{minipage}{0.48\textwidth}
    \includegraphics[width=0.8\textwidth]{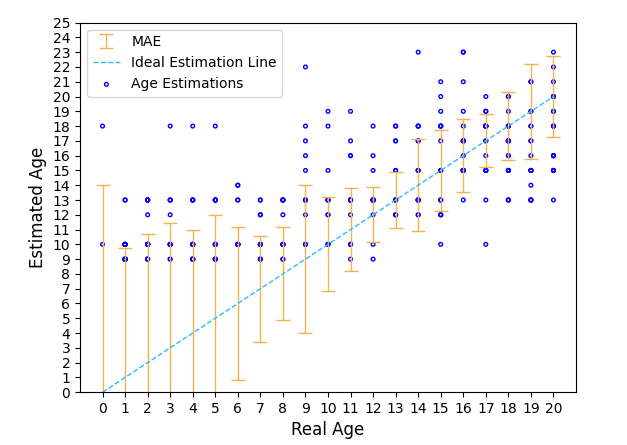}
    \caption{Age estimation results including the Mean Absolute Error.}\label{fig:mae_figure}
  \end{minipage}
  \hfill
  \begin{minipage}{0.48\textwidth}
    \includegraphics[width=0.8\textwidth]{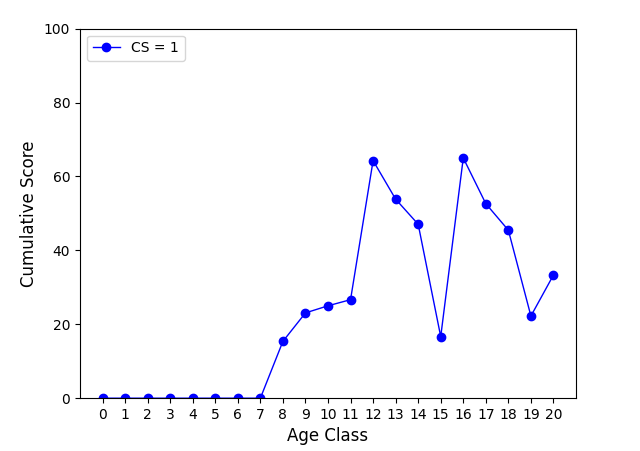}
    \caption{Cumulative Score of 1 per age class.}\label{fig:cs1_figure}
  \end{minipage}
\end{figure*}

Since age estimation needs to be accurate, especially near the age of consent boundary, the researcher decided to define a stringent CS threshold: the two neighbouring age classes. Any estimated result that lies within that range of $\pm$ 1 year, gets counted as an accurate result. The results of the CS are presented in Figure \ref{fig:cs1_figure}. It can be observed that the CS increases with the age of the depicted individual and, therefore, correlates with the MAE. The performance can be summarised as follows: The greater the integral of the CS function over a particular set of age classes, the higher the accuracy within that age range.

\begin{figure*}
  \centering
  \begin{minipage}{0.48\textwidth}
    \includegraphics[width=0.8\textwidth]{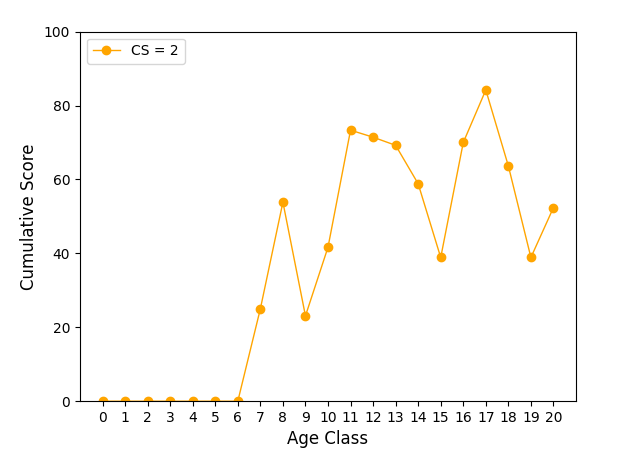}
    \caption{Cumulative Score of 2 per age class.}\label{fig:cs2_figure}
  \end{minipage}
  \hfill
  \begin{minipage}{0.48\textwidth}
    \includegraphics[width=0.8\textwidth]{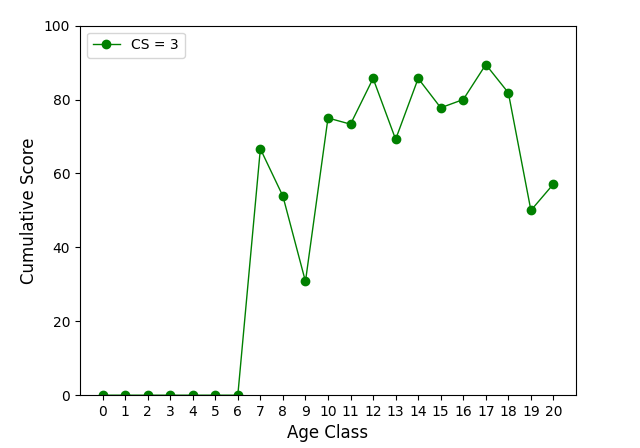}
    \caption{Cumulative Score of 3 per age class.}\label{fig:cs3_figure}
  \end{minipage}
\end{figure*}

The CS has been further divided into two additional sub metrics: CS scores with an interval range of 2 years (Figure \ref{fig:cs2_figure}) and 3 years (Figure \ref{fig:cs3_figure}) respectively.
This approach was inspired by multiple digital forensics experts that have been interviewed in the course of this research (see Discussion and Conclusion). The argument behind that approach is that a single CS cannot show the full picture of classification results. Three different CS intervals provide digital forensic investigators with enhanced intelligence about a set of images depicting the same individual. This can be particularly important when images are imperfect.

Another factor that can negatively impact the CS values is over- and underfitting the model. Since the pre-trained model in this experiment is likely to be overfitted on adults, overfitting can be a contributing factor to low CS values and high MAEs in specific age ranges. Underfitting, on the other hand, would be indicated by a high bias and thus a large difference between training and testing set.

\subsection{Comparison with Recent Results in the Literature}

In comparison, \cite{rothe2016} have used their IMDB-WIKI as a training set and, inter alia, the FG-NET as a testing set. The FG-NET, as well as the MORPH dataset, are used as their standard benchmark. Their DEX method achieved a remarkable MAE of only 3.09 on the FG-NET with fine-tuning and 4.63 without fine-tuning. \cite{wang2015} achieved an MAE of 4.26 on the FG-NET dataset, whereas \cite{chen2013} had attained an MAE of 4.67.

However, since the FG-NET dataset contains 1002 images but only 82 distinct individuals, \cite{chen2013}, \cite{rothe2016} and \cite{wang2015} were effectively testing their model on only 82 individuals. This implies that their testing set is biased on 82 individuals, which result in poor testing quality. In contrast, this research is testing the model with ~327 individuals and with an (almost) evenly distributed age range from 0 to 20- year-olds.

It is also fair to assume that the MAE in the age range [0; 20] was slightly higher, since the estimation results for older individuals are more precise due to the age distribution of the IMDB-WIKI pre-trained model.

The analysis and comparison of computational time is not particularly relevant for this research because an existing pre-trained model was used and the model was run in a VM on the CPU. Other authors have heavily focused on computational time and algorithmic performance \citep{anand2017}, and therefore the results cannot be compared to these publications.

The findings reveal that the age estimation performance is varying across different age groups. In particular, images depicting children and young adolescents are suffering from high MAE and low CS. Although this suggests that the model performance is inadequate, the poor results in this age range can be partly traced back to the overfitting of the facial image dataset of the pre-trained model and the diversity of the testing dataset with regard to image quality, facial rotations and partial occlusions.

The results also show that the age estimation performance is increasing around the age of consent boundary, which is especially important for digital forensic investigations. Some age ranges are even showing lower MAEs compared to the results in recent literature. However, theses MAEs are calculated over an entire dataset and not only a specific age range.

\section{Discussion and Conclusions}\label{Discussion and Conclusions} 

From a quantitative perspective, the findings suggest that even the most diverse datasets can reveal accuracy issues in certain age groups if not enough adequate training input is provided. For instance, since images depicting five-year-old children are very scarce in the used training dataset, the neural network is not able to process this kind of input accordingly. Hence, the achieved results stress the need for more gender and racial diversity as well as an increased image density in the lower age ranges. This study's conclusion agrees with recent and influential literature in the field \citep{bessinger2016, merler2019, ryu2018, ryu2017}.  Since every face is different and reflects variations in, inter alia, race, ethnicity, gender, age and geography, neural networks can be biased in favour of external characteristics. However, the accuracy of a neural network should not vary because of these factors and, therefore, close attention needs to be paid to developing balanced datasets that fairly represent the target group.

While there are differences in the experts' backgrounds and fields, the results of the software evaluation show that all interview participants are satisfied with the efficacy of the prototype in the following areas: 
\begin{enumerate}[noitemsep]
\item Presenting the age value as the key piece of information and supporting it with an age probability distribution graph.
\item Providing additional information about the image itself in the form of metadata.
\item Using three different cumulative scores to determine the efficacy of the age estimation prototype.
\end{enumerate}

The first area is a crucial piece of information since digital forensic investigators do not support the idea of solely presenting crisp age values. An age range is a better expression of accuracy and confidence intervals, which is also consistent with the presentation of results in the literature \citep{anda2018, chen2017, dong2016}. Moreover, an often overlooked aspect is the persuasiveness in court. For instance, if a digital forensic investigator in a child abuse case is presenting evidence in court, they are never just relying on a single age value and, therefore, a DSS such as an age estimation software should present the results in a similar way.

Furthermore, the analysis of the narrative revealed that all four interview participants agree on the provision of additional information in the form of metadata. Most age estimation solutions today are displaying valuable information such as timestamps and file sizes in order to correlate different pieces of evidence with one another.

However, participant A1, A2, and B also point out that a functioning age estimation prototype needs to have the ability to classify a group of images simultaneously and participants B and C stressed the need for an increased age range in the dataset. These suggestions are perfectly reasonable as a DFI usually comprises several gigabytes if not terabytes of image and video material and age classification is required for every single case. Increasing the age ranges of this particular research can be easily achieved by merging additional datasets into the existing one if needed. Since many facial image datasets \citep{bianco2017, chen2014, liu2016, ricanek2006} include a reasonable amount of images depicting individuals between the ages of 20 and 30, existing datasets can be comfortably extended.
    
\subsection{Enhancing Digital Forensic Investigations}

\begin{figure*}
\caption {Interview participants results and suggestions.}
\begin{center}
\includegraphics[width=0.8\textwidth]{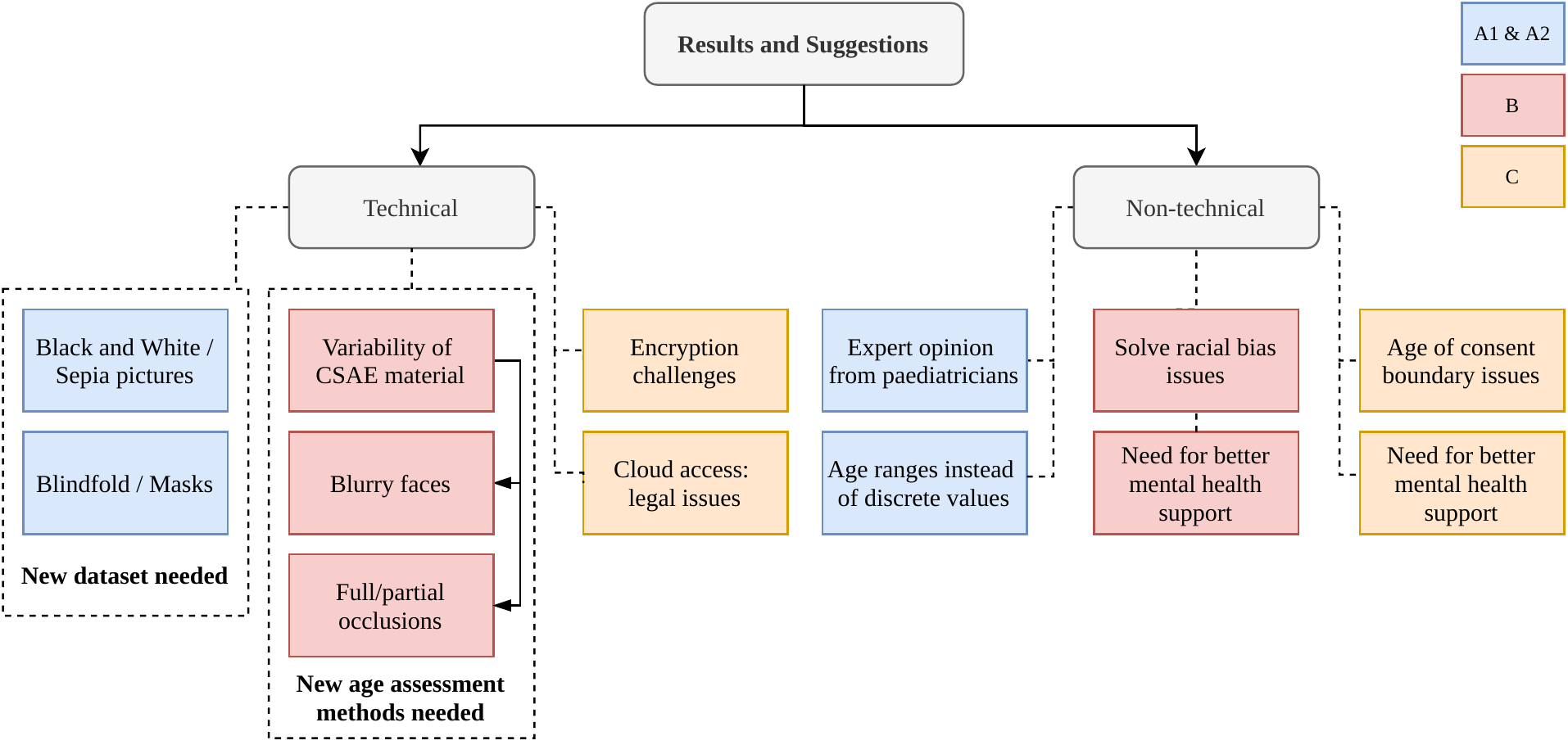} 
\end{center}
\label{fig:results_all_participants}
\end{figure*}

The interview participants A1 and A2 give the most comprehensive feedback on both the technical and non-technical aspects of this research. The main problems they have identified and how these issues can be addressed are presented in Figure \ref{fig:results_all_participants}.

One main suggestion is to include black and white as well as sepia-style images into the dataset since digital forensic investigators are often seizing images with colour filters. However, mixing up coloured and clean-cut images with filtered ones can reverse the positive effects of a racially diverse dataset. By including unnatural images, the neural network could adjust its weights in favour of black and white images or vice versa if they represent a certain proportion of the total images. Referring back to the quantitative findings of Chapter 5, it is evident that the ratio of images in a particular subclass is decisive for the accuracy of the results, regardless of whether it is gender, racial, or age diversity. This argument is also underlined by the research from \cite{zhu2018} who studied the effects of racially balanced datasets and highlighted the importance of maintaining a balance between images depicting black and white, male and female individuals. Hence, in order to cope with black and white/sepia filters, independent datasets must be developed.

Another issue raised by the participants A1, A2 and B is the blindfolding and covering of victim's faces. This enables child abusers to conceal the identity of the victims and possibly stay undetected for a longer period of time because different cases cannot be related to one another based on the child's face. Even unintentional disguises of the face such as blurs and bad image quality, in general, are causing problems for investigators as well as for age estimation software. In order to overcome the problem of partial or full facial concealments, age assessment needs to progress in a different direction. Besides the face of an individual, many other features of the human body are carrying age information. Few studies, including \cite{machado2019}, are exploring age estimation from facial proportions, which still could be effective if victims are blindfolded or wearing masks. However, alternative methods, such as cranial and dental age estimation as well as the age estimation based on the sexual development stages are difficult if not impossible to carry out by only having image and video material available.

As presented in Figure \ref{fig:results_all_participants}, participant B points out similar issues in digital forensic investigations: Besides the massive backlog and racial bias, blurry images (intentional and unintentional face blurs), as well as full/partial face occlusions are causing problems in age estimation. Furthermore, the thematic analysis of interview B and C revealed that there is a need for better mental health counselling.

In contrast, participant A1 and A2 personally pointed out that they have always received adequate mental health counselling and had various other opportunities to cope with work-related psychological stress. In general, however, the data found in \cite{seigfried2018} suggests that the majority of individuals performing duties related to CSAM are not adequately coping with work-related stress, based on a sample of 129 law enforcement officers. \cite{carleton2018} support this view, who discovered that mental disorder symptoms are more prevalent within public safety personnel than previously expected. Hence, since human investigators will remain a vital part of digital forensic investigations for the next decades, much more attention needs to be paid to mental health counselling.

Participant C addressed the questions from a more generic point of view by highlighting the most prevalent challenges in the digital forensics domain – the encryption and obfuscation of CSAM (Figure \ref{fig:results_all_participants}). This view is underlined by the findings of the Literature Review on the massive growth of cloud forensics as well as the increase in encrypted image and video material. 
Evidence collection and recovery is challenging within a cloud environment, not only because of the sheer amount of data but also due to jurisdiction problems that can occur \citep{pichan2015}.

From a non-technical point of view, the interview with participant C revealed that the age of consent boundary needs to be more in the focus of researchers. Although the emphasis on the exact age estimation of a victim remains important, the binary categorisation into adult or non-adult should be the main target. This research already addresses this problem with the embedded age estimation functionality.

\subsection{Limitations and Future Work}

The confirmability and transferability of qualitative research could be further enhanced by testing the prototype efficacy more broadly. Currently, this research is solely aimed at studying digital forensic age estimation in the UK. However, processes vary significantly across different countries and legislations. To get a more profound picture of the potential efficacy of the research, the age estimation prototype needs to be tested in cooperation with multiple digital forensic departments across different countries. Hence, future research could overcome this problem by extending the qualitative interviews and software evaluation to a survey with a larger sample size.

In future work, the testing dataset should be enlarged to get more accurate insights into the underlying dynamics of various pre-trained age estimation models. New datasets can be collected that particularly focus on black \& white images as well as partial facial concealments. Unfortunately, since many well-suited facial image datasets remain private, they cannot be merged with the new dataset. The increase in demand for publicly-shared datasets is not only limited to this domain but can be observed across many other scientific fields. In order to maximise the knowledge gain by researchers, recent studies \citep{patel2017, shrestha2018} are reviewing options to apply blockchain technology to the secure distribution of scientific datasets.

Furthermore, there is remaining opportunity for further research in the area of coalitions against CSAE. A promising approach is presented by The Technology Coalition (\citeyear{techcoaltion2020}). This organisation brings together different companies across the technology industry, such as Microsoft, Google and Facebook, and relies on sharing best practices and coordinating efforts to combat CSAE. In the last decade, these companies have developed and consistently supported the research in the area of machine learning. By leveraging their seemingly unlimited access to knowledge, computing power and other resources, big coalitions are the most promising way to combat CSAE. Hence, researchers in the area of age estimation should be on the lookout for ways to integrate their work and thus contribute to existing coalitions in order to create more accurate solutions.

\subsection{Conclusion}

The aim of this research is to enhance digital forensic investigations by assessing current age estimation solutions, reviewing facial image datasets and developing an age estimation prototype. The results show that even very large publicly available facial image datasets and age estimation models can be unsuitable for digital forensic investigations due to factors such as image scarcity in lower age ranges and low image diversity.

By categorising, evaluating and critiquing the age estimation challenges as well as the surrounding issues in the domain, this research shows that digital forensic investigations can benefit from age estimation software by reducing the mental stress on investigators and potential cognitive bias.

Based on the quantitative analysis of the age estimation results, it can be concluded that estimation results around the age of consent boundary show comparable performance to state-of-the-art solutions. Although the age estimation accuracy of young children is imprecise, it does not greatly affect DFIs since the approximate age of young children can generally be correctly determined by investigators.

The qualitative interview and software evaluation support this research with qualitative insights from digital forensic experts. They aid the presentation of age estimation results with age probability graphs and the accuracy evaluation using the CS. The findings also reveal that investigators are concerned about racial bias, blurry images as well as full/partial face occlusions.

Returning to the original problem statement, this research further progresses digital forensic investigations towards fully-automated age estimation processes by showing the immense potential of CNNs and highlighting the limitations of existing solutions. Research that is built upon this thesis can enhance workflows in DFIs, reduce the psychological burden on investigators and thus ultimately combat CSAE-related cybercrime.






\footnotesize
\singlespacing
\bibliographystyle{agsm}

\bibliography{bibliography.bib}
\normalsize






\end{document}